\documentclass[letterpaper, 10 pt, conference]{ieeeconf}  

\IEEEoverridecommandlockouts                              
\usepackage{graphics} 
\usepackage{epsfig} 
\usepackage{mathptmx} 
\usepackage{times} 
\usepackage{amsmath} 
\usepackage{amssymb}  
\usepackage{multirow}
\usepackage{bm}
\usepackage{xcolor}
\usepackage{ifthen}
\usepackage[capitalise,nameinlink]{cleveref}
  \crefname{section}{Sect.}{Sect.}
  \Crefname{section}{Section}{Sections}
  \crefname{figure}{Fig.}{Fig.}
  \Crefname{figure}{Figure}{Figures}
  \crefname{table}{Tabl.}{Tabl.}
  \Crefname{table}{Table}{Tables}
\newtheorem{definition}{Definition}

\ifthenelse{\boolean{false}}{
\definecolor{higashi}{rgb}{0.2,0.4,0.6} 
\definecolor{cristian}{rgb}{0.3,0.6,0.3} 
}{
\definecolor{higashi}{rgb}{0,0,0} 
\definecolor{cristian}{rgb}{0,0,0} 
}
\title{\LARGE \bf
Functionally Divided Manipulation Synergy\\ for Controlling Multi-fingered Hands
}

\author{Kazuki Higashi$^{{1}{*}}$, Keisuke Koyama$^1$, Ryuta Ozawa$^{2}$, Kazuyuki Nagata$^{3}$, Weiwei Wan$^{1,3}$ and Kensuke Harada$^{1,3}$
\thanks{*This work was supported in part by the NACHI-FUJIKOSHI Corporation.}
\thanks{$^{1}$Graduate School
of Engineering Science, Osaka University; 1-3 Machikaneyama-
cho, Toyonaka, Osaka 560-8531, JAPAN}
\thanks{$^{2}$School of Science and Technology,
Meiji University}%
\thanks{$^{3}$National Institute of Advanced
Industrial Science and Technology}%
\thanks{$^{*}$ Corresponding author e-mail: \newline \hspace*{10mm}
        {\tt\small higashi[at]hlab.sys.es.osaka-u.ac.jp}}%
}

\begin{document}

\maketitle
\thispagestyle{empty}
\pagestyle{empty}

\begin{abstract}
\color{cristian}
Synergy supplies a practical approach for expressing various postures of a multi-fingered hand. However, a conventional synergy defined for reproducing grasping postures cannot perform general-purpose tasks expected for a multi-fingered hand. 
Locking the position of particular fingers is essential for a multi-fingered hand to manipulate an object. When using conventional synergy based control to manipulate an object, which requires locking some fingers, the coordination of joints is heavily restricted, decreasing the dexterity of the hand. 
We propose a \textit{functionally divided manipulation synergy} (FDMS) method, which provides a synergy-based control to achieves both dimensionality reduction and in-hand manipulation. 
In FDMS, first, we define the function of each finger of the hand as either ``manipulation" or ``fixed." Then, we apply synergy control only to the fingers having the manipulation function, so that dexterous manipulations can be realized with few control inputs.
The effectiveness of our proposed approach is experimentally verified. 

\end{abstract}


\color{cristian}

\section{Introduction}
\label{sec:introduction}
Synergy provides a method to simultaneously control multiple Degrees of Freedom (DoFs) of a multi-fingered hand to produce coordinated movements from a limited set of input signals. One of the common strategies for controlling multi-fingered hands is to adopt synergy for their control inputs \cite{Pisa_SoftHand2, unipi_hand, sma_array, X_Hand, dexmart, TSS, mechanical_selector, mapping, extracting_synergy, teleop, impedance_control_synergy, prof.ficuciello1, prof.ficuciello2, prof.ficuciello3}.
For grasping an object, a synergy, hereafter called grasp synergy, can be constructed by applying the principal component analysis (PCA) to the human's hand grasping postures. The grasp synergy is then used to express various postures with the multi-fingered hand using a few principal components \cite{hand_synergies,soft-tissue,hand_analysis}. 

The main purpose of a multi-fingered hand is to manipulate objects, including the use of tools. Using conventional synergy-based control, object grasping \cite{prof.ficuciello1} and simple in-hand manipulation (changing the pose of an object) \cite{Pisa_SoftHand2} have been realized. Nevertheless, it is very challenging to achieve the required posture expressions for using human tools with the framework of conventional synergy-based control. Therefore, we focus on the manipulation of human tools with a new synergy-based control method.

Kamakura \cite{kamakura_en} measured and analyzed human hand movements when manipulating an object. Kamakura's research showed that in 45.5\% of human hand motions observed in daily life, the position of particular fingers is fixed in place. According to this observation, we hypothesis that to improve control's performance of a multi-fingered hand during in-hand manipulation tasks, the position of specific fingers needs to be fixed. The fingers that should hold a fix position are, for example, those that are not needed for the manipulation task or those that are needed to hold the position of a grasped object.

However, the grasp synergy-based control assumes the coordination of all joints of a multi-fingered hand in order to express human-like grasping postures. Thus, fixing the position of specific fingers heavily limits the control space of the grasp synergy. The precise posture expression, with reduced dimensionality, required for object manipulation is difficult to realize by using conventional grasp synergy-based control.

\begin{figure}[t]
  \begin{center}
    \includegraphics[width=\linewidth]{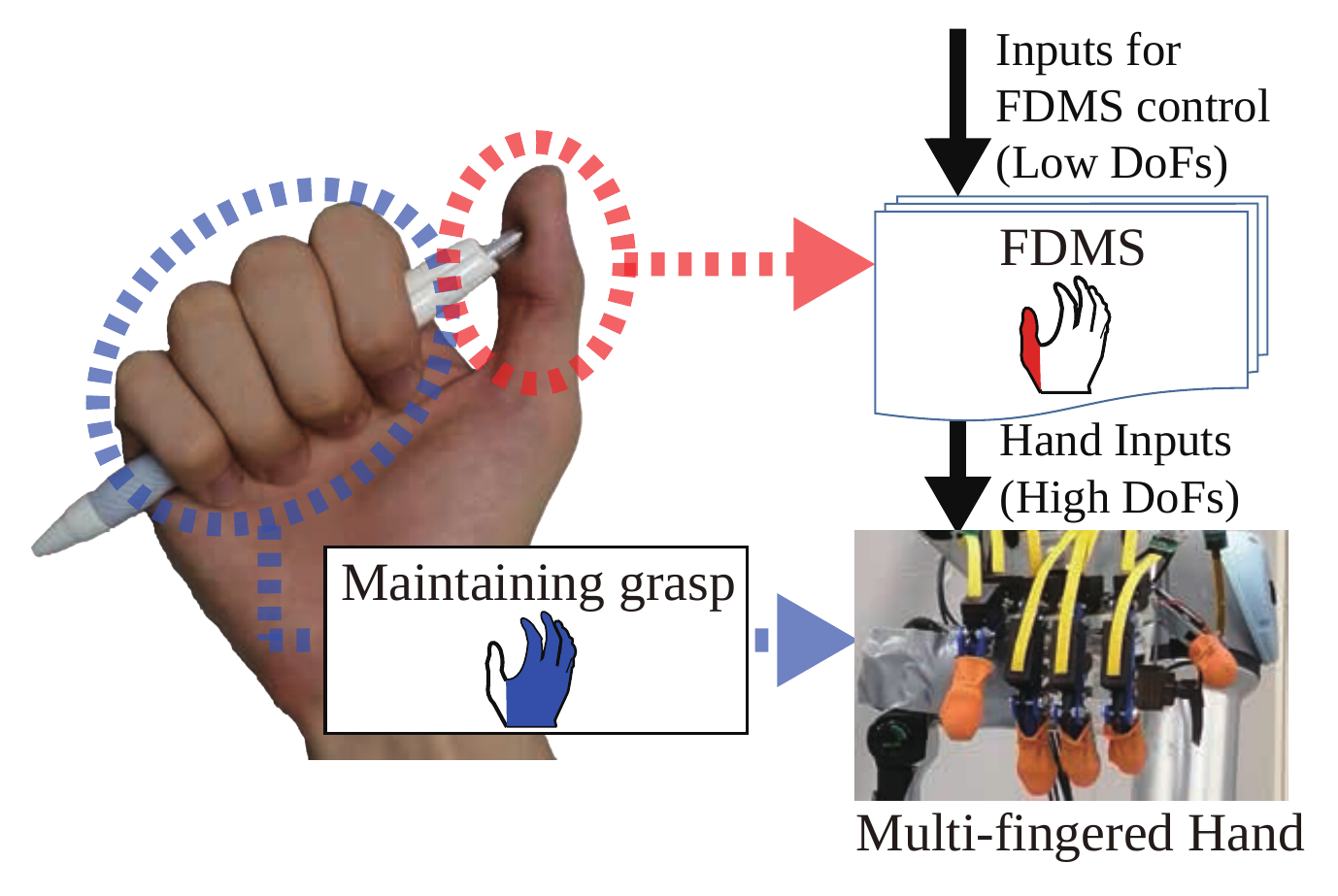}
    \caption{Overview of our proposed method. Our FDMS control method consists of the differentiation of the function of each finger of the hand; the ones needed for manipulation and the ones that do not need to be moved (fixed).}
    \label{fig:summary_of_fdms}
  \end{center}
\end{figure}
Considering the above mentioned problem, this paper proposes a \textit{Functionally Divided Manipulation Synergy} (FDMS) method, which provides a synergy-based control for a multi-fingered hand when manipulating an object, in which some of the fingers are used to manipulate an object while others are fixed in place. 
\Cref{fig:summary_of_fdms} shows an overview of the proposed FDMS. Said figure shows an example of clicking downward on the top of a pen with one's thumb. In this example, the position of index, middle, ring and little fingers are kept fixed to maintain the grasp of the pen. 
First, various grasping postures of a human hand are measured in advance. Then, we construct FDMSs based on the principal component vectors of the measured human hand's postures including only the fingers or joints needed for the manipulation. 

Similarly, we consider the situation where a robot first picks a pen and then press the top of the pen. The robot first grasps the pen by using the grasp synergy based control, then uses the FDMS method to control only the thumb to click downward the top of the pen while other fingers are used to maintain a grasp of the pen. 
In the said case, we can switch among multiple FDMSs and fix different fingers according to the task. Therefore, we propose a framework named the Switching Synergy Framework. The Switching Synergy Framework makes the FDMSs useful to perform various sequential tasks. 

The main contribution of our work is to provide a synergy control framework for performing a task with a few dimensional control inputs. For this purpose, we introduce the FDMS method to assign a function to each finger (Manipulation or Fixed). FDMS-based control is found out to achieve a higher task success rate than conventional grasp synergy-based control. 
For comparison, we also introduce the task-specific synergy composed only of a set of hand postures of performing a specific task. 



This paper is organized as follows. The related work on the synergy and multi-fingered hands are introduced in \Cref{sec:related_work}. Our proposed method using FDMS is depicted in \Cref{sec:synergy} and \Cref{sec:fdms}. Our proposed Switching Synergy Framework which make FDMS useful to apply in sequencial tasks is described in \Cref{sec:ssf}. \Cref{sec:experiments} describes the experiments conducted to evaluate the proposed methods. The results and comparisons between methods are described in \Cref{sec:results}. 

\color{black}

\section{related work}
\label{sec:related_work}
\color{higashi}

The concept of synergy has drawn attention in the robotics research community. Many researchers have researched it since it provides an effective method for controlling multi-fingered hands with a few dimensional control inputs. 
Santello et al. \cite{hand_synergies} measured human hand motions during daily life, assuming a 15 DoFs kinematics model and showed that 80\% of the movements could be realized with just two principal components. 
Similarly, Todorov et al. \cite{complicated_synergy} showed that four principal components represent 85\% of the movements when assuming a 20 DoFs kinematics model. 
Xiong et al. \cite{X_Hand} developed a synergy-based multi-fingered hand which can realize 30 kinds of grasping poses out of 33 poses defined by Feix \cite{taxonomy}. 
Chen et al. \cite{Yue1} developed a synergy-based hand based on the 2 DoFs angular velocity synergy, implemented by using cam and underactuation mechanism.

Although the grasp synergy provides an effective method for controlling multi-fingered hands, the research on performing a task with synergy-based control is limited.  

Kent et al. \cite{TSS} proposed the task dependent temporally synchronized synergies (TSS) for performing a task by using a prosthetic hand according to the EMG signal of a human. 
TSS is an attractive method that enables us to generate dexterous motion of an object, but to create a TSS, humans need to measure the motion for each task. In general, a multi-fingered hand is required in a large number of tasks, so the TSS method has a very high measurement cost. 
Our method, FDMS, is constructed from a variety of grasping postures, similar to grasp synergies. Thus, the measurement cost is low, and it is only necessary to assign specific functions to each finger to perform tasks.

Santina et al. \cite{Pisa_SoftHand2} developed a synergy-based multi-fingered hand and demonstrated the effectiveness of the adaptive synergy \cite{unipi_hand} through some complex tasks such as grasping and opening a screw cap.
However, the conventional grasp synergy-based control is not suitable for performing a task, since if we try to fix the position of a finger of multi-fingered hand when performing a task, the control space spanned by its principal component vectors is heavily restricted and cannot express postures required for the advanced tasks. Therefore this property makes the grasp synergy poor to perform manipulation tasks.
On the other hand, our proposed synergy-based control can perform dexterous tasks such as manipulating an object while grasping that without being affected by fixed fingers because we consider only fingers assigned the manipulating function.

\section{synergy}
\label{sec:synergy}
\color{cristian}
\subsection{synergy control}
\label{ssec:synergy_control}

Synergy, in the context of control of a multi-fingered hand, is defined to be the coordination of joints and defines the basis of the joint-angle space. In this section, we first formulate synergy introduced in this research. 

By using a set of hand poses, the posture sequence $\bm{X}$ used for formulating synergy can be defined as:
\begin{equation}
  \label{eq:data}
  \bm{X} = [ \bm{p}_1, \bm{p}_2, \cdots, \bm{p}_j, \cdots, \bm{p}_n ]^{T},
\end{equation}
where $\bm{p}_i$ ($i=1,\cdots,n$) denotes a $d$-dimensional vector expressing a configuration of a multi-fingered hand. 
We assume that each pose included in $\bm{X}$ is sufficiently diverse. 



In this research, we make a multi-fingered hand manipulating an object by using some of the fingers while others are used to grasp or simply fixed. We formulate such division of finger functions.  
By defining the indices of all joints included in the hand as $J = \{1, 2, ... , d\}$, we can define the subset of $f$ joints from $J$ as $J^{\mathrm{sub}} = \{ s_1, \cdots, s_j, \cdots, s_f \} \subseteq J$ where $J^{\mathrm{sub}}$ is defined to be a set of joints included in some of the fingers to realize a specific function. 
By using $J^{\mathrm{sub}}$, a hand pose $\bm{p}_{j}^{\mathrm{sub}}$ can be represented as:
 
\begin{equation}
  \label{eq:extracted_joints}
  \bm{p}_i^{\mathrm{sub}} = [ p_{i, s_1}, p_{i, s_2}, \cdots, p_{i, s_j}, \cdots, p_{i, s_f}]^T
\end{equation}

\noindent
From \eqref{eq:extracted_joints}, we can obtain the posture sequence $\bm{X}^{\mathrm{sub}}$ consisting of $\bm{p}_{i}^{\mathrm{sub}}, i=1,\cdots,n$. 
The principal component analysis (PCA) of the extracted hand pose is equivalent to the eigenvalue problem for its covariance matrix. The eigenvectors $\bm{a}_1, \cdots, \bm{a}_{f}$ and the eigenvalues $\lambda_1, \cdots, \lambda_{f}$ represent the coordination of joints and the variance of each principal component, respectively. Finally, we can define a synergy matrix $S$ for the $n_s$-dimensional synergy control as follows:
\begin{equation}
    \label{eq:synergy_matrix}
    S = [\bm{a}_1, \cdots, \bm{a}_{n_s}]
\end{equation}

\subsection{synergy definition}
\label{ssec:synergy_definition}
Now we define three different synergies, the most commonly used grasp synergy, the functionally divided manipulation synergy proposed in this paper and the task-specific synergy. 

\noindent
We first introduce the grasp synergy which is equivalent to the conventional definition of synergy:

\begin{definition}{\bf Grasp Synergy}\ \ 
    is defined to be a synergy composed of various grasping postures in $\bm{X}$. Grasp synergy considers the total joint space $J$ of a human hand or a multi-fingered hand. 
\end{definition}

\noindent
On the other hand, we aim to make a multi-fingered hand to perform a task. We introduce the following two definitions on synergy for performing a task. Among the two definitions, the Functionally Divided Manipulation Synergy (FDMS) proposed in this paper will be define later in the next section. For the comparison study, we introduce synergy for performing a specific task:

\begin{definition}{\bf Task-Specific Synergy}\ \ 
    of the $k$-th task is defined to be a synergy composed of hand postures $\bm{X}^T_k$ performing the $k$-th task considering the total joint space $J$ of a human hand or a multi-fingered hand. The synergy matrix of the $k$-th task-specific synergy is denoted as $S^T_k$.
\end{definition}

\subsection{postures approximation}
\label{ssec:postures_approximation}
In synergy-based control, it is important to evaluate how precisely a hand posture can be realized for performing a requited task. To evaluate the capability to express such function-required postures, we verify whether or not a task succeeds when a hand postures are presented in the synergy space. By using the synergy matrix, an approximated posture $\hat{\bm{p}}$ is given by:

\begin{equation}
\label{eqn:approxed_posture}
    \hat{\bm{p}} = SS^T\bm{p},
\end{equation}

\noindent
where $SS^T$ denotes an orthogonal projection matrix projecting a joint angle vector to the synergy space. This accuracy of approximated posture is used to evaluate each synergy. The posture sequence $\hat{\bm{X}}$ approximated by a synergy can be expressed by:

\begin{equation}
\label{eqn:approxed_dataset}
    \hat{\bm{X}}^T = SS^T\bm{X}^T = [\hat{\bm{p}}_1, \hat{\bm{p}}_2, \cdots , \hat{\bm{p}}_j, \cdots, \hat{\bm{p}}_N]
\end{equation}

\section{Functionally Divided Manipulation Synergy (FDMS)}
\label{sec:fdms}



In this section, we further formulate FDMS method based on the typical movement units of human hands.
We assign two kinds of functions (Manipulating and fixed) to each finger based on a study of finger motion analysis for manipulating an object \cite{kamakura_en}. We also identify necessary FDMSs based on typical finger movements and its frequency. 

\subsection{Typical unit movements of human hands}
\label{ssec:typical_motion}
\begin{figure*}[t]
  \begin{center}
    \includegraphics[width=0.95\linewidth]{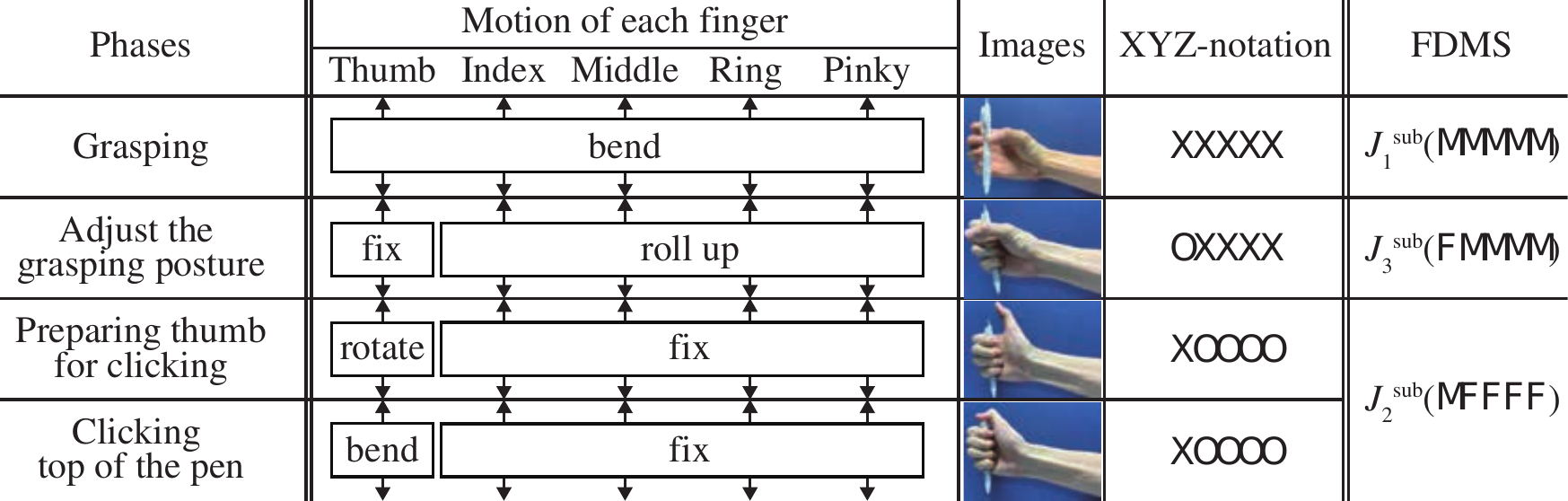}
    \caption{Appearance of a pen-knocking task. (Phases): describes the purpose of each unit movement. (Motion of each finger): shows the kind of finger motion and the relation between motion of each finger. (Images): depicts each phase of the task. (XYZ-notation): shows the notation of each movement by using XYZ-notation described in \Cref{ssec:typical_motion}. (FDMS): shows what kind of FDMS defined in \Cref{ssec:listing_fdms} use for the movement}
    \label{fig:task_example}
  \end{center}
\end{figure*}

To analyze the human hand motion, we introduce the Kamakura's XYZ-notation, which defines the finger movement when manipulating an object \cite{kamakura_en}. The XYZ-notation method decompose a series of human hand motions, manipulating an object, into movement units. The movement unit for a given manipulation task is defined with respect to the the movement of each finger, considering them in sequential order from the thumb to the little finger. Each finger is assigned a symbol \texttt{X}, \texttt{Y}, \texttt{Z}, or \texttt{O} according to the following conditions. Kamakura defined discretized cluster of a finger movement. If a finger movement is categorized into the same cluster, we assign the same symbol. 
Firstly, if there is any movement of the thumb, assign the symbol \texttt{X}. Next, if the movement of the index finger is the same as the movement of the thumb, assign the same symbol \texttt{X}. If the movement is different, assign the symbol \texttt{Y}. Similarly, if the movement of the middle finger is same as the thumb, assign the symbol \texttt{X}, if different from the thumb but the same as the index finger, assign the symbol \texttt{Y}. Moreover, if the movement of the middle finger is different from both the thumb and the index finger, assign it the symbol \texttt{Z}. The symbol \texttt{O} is assign to a finger that has no motion during a given manipulation task. According to \cite{kamakura_en}, for most of daily tasks, each finger performs at most three different motions. This is because only XYZ are used to express finger motion. According to the XYZ-notation, movement of a human hand can be described with five characters. An example of using the XYZ-notation to characterize a task is shown in \Cref{fig:task_example}. In said example, the task of manipulating a pen is considered as being composed of four phases; grasping the pen, preparing grasping posture for clicking the top of the pen, preparing posture of the thumb, and then clicking downward the top of the pen, to expose the ballpoint.

\begin{table}[t]
\centering
\caption{Typical movement unit and its frequency\cite{kamakura_en}}
\label{tab:kamakura}
\begin{tabular}{c||r}
Movement unit& Frequency[\%] \\ \hline 
\texttt{XXXXX} & 8.5 \\
\texttt{XYYYY} & 9.0 \\
\texttt{XOOOO} & 8.9 \\
\texttt{OXXXX} & 3.5 \\
\texttt{XXYYY} & 1.9 \\
\texttt{XXOOO} & 3.2 \\
\texttt{OOXXX} & 5.1 \\
\texttt{OOOXX} & 2.4 \\
\texttt{XYZZZ} & 3.5 \\
\texttt{XYOOO} & 4.8 \\
\texttt{OXYYY} & 2.6 \\
\texttt{OXOOO} & 8.9 \\
\texttt{XYYOO} & 1.9 \\
\texttt{OOXOO} & 1.9 \\
\texttt{OOXXO} & 1.3 \\
\texttt{XYYYO} & 1.0 \\ \hline 
Total & 68.4 \\ 
\end{tabular}
\end{table}

\Cref{tab:kamakura} shows the typical unit movements and their frequency obtained from the analysis of human motions manipulating an object \cite{kamakura_en}. Said table lists all the unit movements whose frequency exceeds 1\%. More than 68.4\% of unit movements can be expressed by using the mentioned four symbols of the XYZ-notation. In addition, unit movements in a state where at least one finger does not move is 45.5\%, implying that it is necessary to fix particular fingers when manipulation an object with human hands. 


\subsection{Necessary FDMSs}
\label{ssec:listing_fdms}

As mentioned above, the XYZ-notation method decompose a series of human-hand motions into movement units. However, each symbol in the XYZ-notation does not explain what function each finger has. In this study, we define the following two symbols to describe the function of each finger, in relation to the XYZ-notation.

\begin{enumerate}
    \item Manipulation : \texttt{M} \\
        is assigned to the fingers required for manipulating an object. This corresponds to \texttt{X}, \texttt{Y}, and \texttt{Z} in the XYZ-notation.
    \item Fixed : \texttt{F} \\
        is assigned to a finger that needs to be fixed with its current posture, either because it is necessary to keep that posture (e.g. to maintain a grasping posture), or because the finger is not necessary for the task. This corresponds to \texttt{O} in the XYZ-notation.
\end{enumerate}

 
To avoid assigning multiple manipulating functions to different fingers within a single movement unit, we divide the movement unit into several units to include only a single manipulating function per movement unit. For example, when we use the computer mouse with a unit movement \texttt{OXYOO}, the index finger is assigned the function of left-clicking and the middle finger is assigned the function of right-clicking. In said case, we divide the movement unit into two units as \texttt{OXYOO} $\rightarrow$ \texttt{OXOOO} + \texttt{OOXOO}. We applied the same operation to all movement units in the \Cref{tab:kamakura}. The resulting movement units are shown in \Cref{tab:listup_fdsyn}.


\begin{table}[t]
\centering
\caption{Units of FDMS (\texttt{M}: A finger included in FDMS construction)}
\label{tab:listup_fdsyn}
\begin{tabular}{l||c}
Joint Subspace & FDMS Units \\ \hline
$J^{\mathrm{sub}}_1$ & \texttt{MMMMM} \\
$J^{\mathrm{sub}}_2$ & \texttt{MFFFF} \\
$J^{\mathrm{sub}}_3$ & \texttt{FMMMM} \\
$J^{\mathrm{sub}}_4$ & \texttt{MMFFF} \\
$J^{\mathrm{sub}}_5$ & \texttt{FFMMM} \\
$J^{\mathrm{sub}}_6$ & \texttt{FFFMM}  \\
$J^{\mathrm{sub}}_7$ & \texttt{FMFFF} \\
$J^{\mathrm{sub}}_8$ & \texttt{MMMFF} \\
$J^{\mathrm{sub}}_9$ & \texttt{FFMFF} \\
$J^{\mathrm{sub}}_{10}$  & \texttt{FFMMF} \\
$J^{\mathrm{sub}}_{11}$ & \texttt{MMMMF} \\
$J^{\mathrm{sub}}_{12}$ & \texttt{FMMMF}  \\
\end{tabular}
\end{table}

Finally, we can define 12 joint subspaces $J^{\mathrm{sub}}_l (l=1, \cdots, 12)$ of FDMS consisting of only fingers assigned \texttt{M} in \Cref{tab:listup_fdsyn}. 
Base on these joint subspaces, we define the FDMS as follows:

\begin{definition}{\bf Functionally Divided Manipulation Synergy (FDMS)}\ \
    is defined to be a synergy composed of various grasp postures $\bm{X}^g$ considering $l$-th joint subspace $J^{\mathrm{sub}}_l$ of a human hand or a multi-fingered hand. The synergy matrix of the FDMS is denoted as $S^f_l$.
\end{definition}

FDMS are constructed in a similar fashion as grasp synergies. Therefore, the FDMS consisting of all joint subspaces is equivalent to the grasp synergy.
\color{black}

\section{Switching Synergy Framework}
\label{sec:ssf}
\color{higashi}

\begin{figure}[t]
  \begin{center}
    \includegraphics[width=\linewidth]{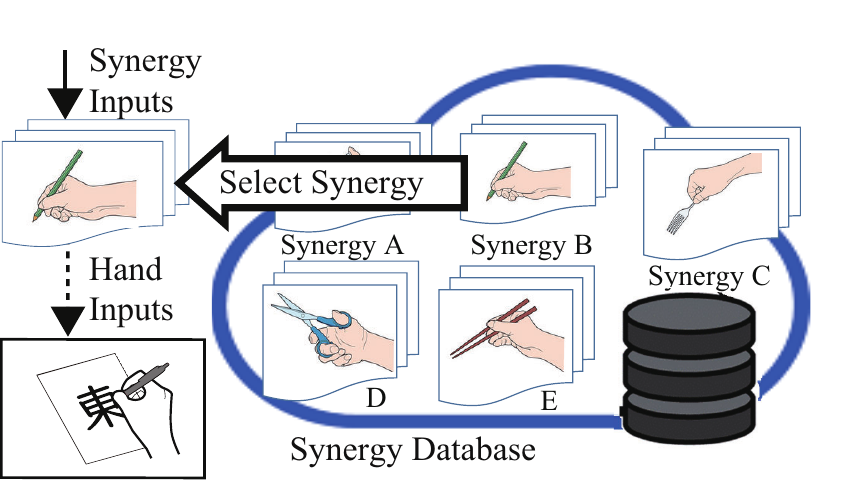}
    \caption{The overview of Switching Synergy Framework. In this research, FDMSs are registered in a synergy database, and the synergies are changed according to the change of functions.}
    \label{fig:ssf}
  \end{center}
\end{figure}


When performing a task, the function required for a finger changes according to the situation. Hence, it is necessary to switch among multiple FDMS according to the change of functions. In this research, we propose Synergy Switching Framework as a framework for switching synergies. \Cref{fig:ssf} depicts the overview of the Synergy Switching Framework. Within said framework, several synergies that can perform various tasks are prepared in advance and a synergy database is constructed. The user inputs the task to be performed into the system, and selects a synergy suitable for performing the task. Depending on the characteristics of the synergies registered in the synergy database, the granularity of the task division and the required inputs vary. In this study, we register 12 FDMSs defined in \Cref{ssec:listing_fdms} to the synergy database.


%

\section{experimental environment}
\label{sec:experiments}
\color{higashi}
\subsection{Human's Hand Pose Measurement}

To measure the posture of the human hand, we use CyberGlove I\hspace{-.1em}I\hspace{-.1em}I \cite{cyberglove3}. The CyberGlove I\hspace{-.1em}I\hspace{-.1em}I can measure the joint angles of the human hand at 100Hz. In this study, only the motion of fingers was considered, without considering the motion of the wrist. Therefore, the angles of 20 joints were measured. 

\subsection{Teleoperating System}

\begin{figure}[t]
  \begin{center}
    \includegraphics[width=0.9\linewidth]{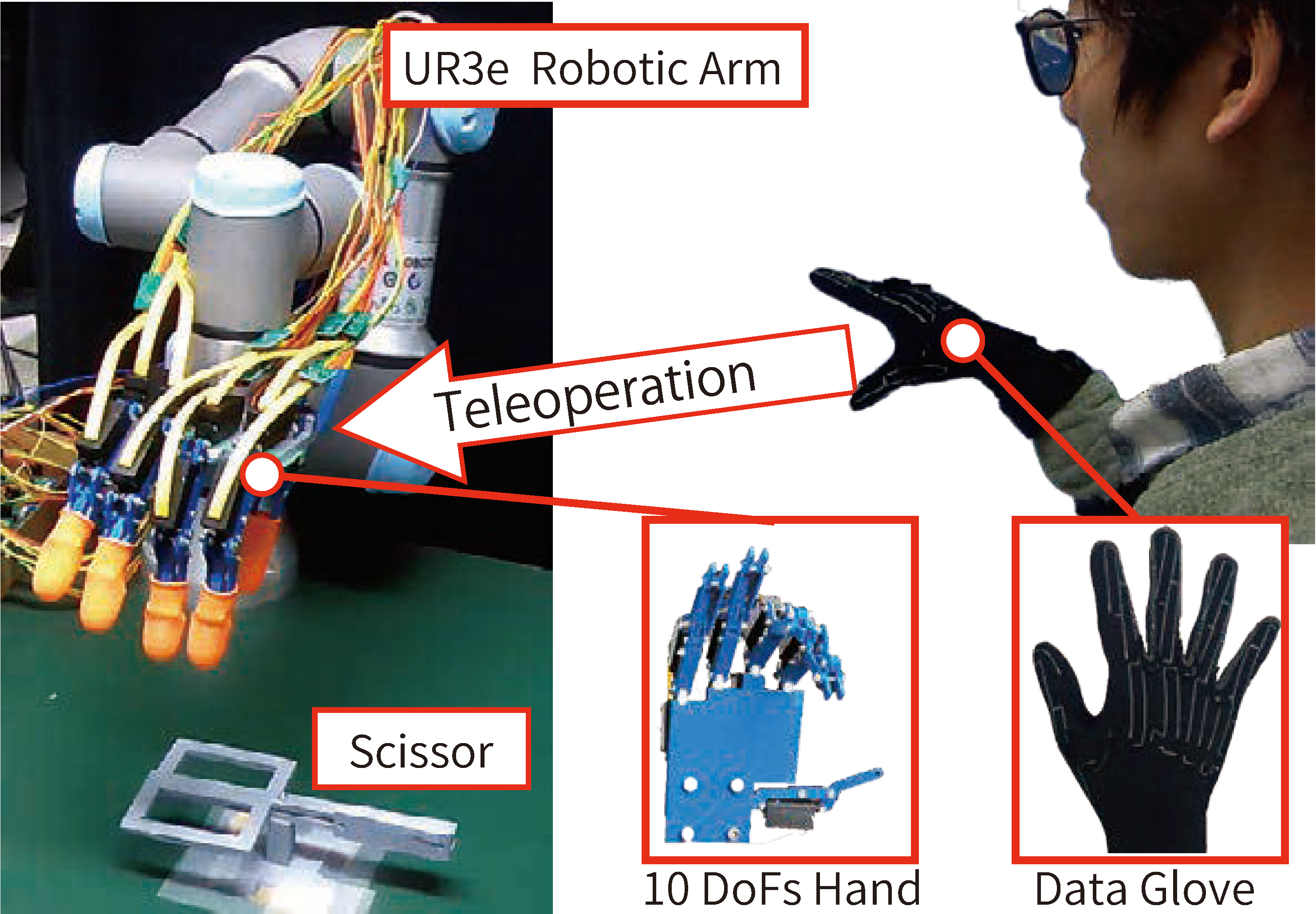}
    \caption{Overview of experimental setup. Teleoperating a 10 DoFs Humanoid Hand using a CyberGlove I\hspace{-.1em}I\hspace{-.1em}I}
    \label{fig:real_environment}
  \end{center}
\end{figure}

The teleoperating system consists of a UR3e robotic arm, a 10 DoFs humanoid hand, and a CyberGlove I\hspace{-.1em}I\hspace{-.1em}I. The subject equipped with the CyberGlove I\hspace{-.1em}I\hspace{-.1em}I performs a task by transferring postures to the 10 DoFs humanoid hand. The movement of the UR3e robotic arm was controlled through keyboard inputs. 

\subsection{Tasks}

In order to evaluate the performance of our method, FDMS, two tasks were conducted: an opening/closing task of scissors and a pushing task of a switch. These tasks utilize tools that have not been thoroughly considered in previous research. In addition, the tasks of using scissors and pressing a switch were selected as a representative example of tasks that required both to grasp and manipulate at the same time.
Following the notation described in \Cref{ssec:listing_fdms}, the mentioned tasks are considered as follows: when using scissors, the thumb, index, and middle finger are thought to have a Manipulation function. For pushing the switch, at first, all fingers are assigned the Manipulation function. Then, only the thumb is assigned with the Manipulation function while the other fingers are assigned the Fixed function.

The overall experimental setup for the tasks described above is shown in \Cref{fig:real_environment}. In both cases, first, the target object is grasped from a fixed initial position on the table. The first task is considered completed if fingers of a humanoid hand inserted into the scissors, and then the scissors were opened/closed three times. Similarly, for the second task, it is considered successful if the switch was fully pressed.

The human operator performed each task for 20 times and measured postures sequences of the hand with the CyberGlove I\hspace{-.1em}I\hspace{-.1em}I. For each task, 10 postures sequences $\bm{X}^T_k$ were used to construct the task-specific synergy and the remaining 10 sequences $\bm{X}^E_m(m=1\cdots 10)$ are used for evaluation.
\subsection{Synergies}
\label{ssec:exprimental_synergies}
We first constructed the grasp synergy from principal components of a variety of grasping postures. The major 33 grasping postures of a human were used based on the grasp taxonomy proposed in \cite{taxonomy}. The humanoid hand postures are measured while the fingers are moved from a parallel and flatten posture to the target grasping posture. Then, the posture sequence $X^g$ was constructed, as mentioned in Section \ref{ssec:synergy_definition}, with postures excluding the overlapping postures. Subsequently, the grasp synergy $S^g$ has been constructed based on the grasp posture sequence $X^g$. 

For comparative evaluation, the \textit{task-specific synergy} was introduced. The task-specific synergy was made of multi-fingered hand postures extracted while performing each specific task, e.g., opening/closing a scissors and pressing a switch, independently. The hand poses were extracted by teleoperating the 10 DoFs humanoid hand. In each case, the task was performed ten times. Then, the \textit{task-specific synergies} were created by principal component analysis as $S^T_{scissors}$ and $S^T_{switch}$.

\begin{table}[]
\centering
\caption{The functions of each finger when performing the scissors-task and the switch-task.}
\label{tab:selected_joints}
\begin{tabular}{l||c|c}

\multirow{2}{*}{Joint} & \multicolumn{2}{c}{Function} \\ \cline{2-3} 
 & \begin{tabular}[c]{@{}l@{}}Scissors-task\end{tabular} & \begin{tabular}[c]{@{}l@{}}Switch-task\end{tabular} \\ \hline
Thumb rot. & \texttt{M} & \texttt{M} \\ \hline
Thumb MCP & \texttt{M} & \texttt{M} \\ \hline
Index MCP & \texttt{M} & \texttt{F} \\ \hline
Index PIP& \texttt{M} & \texttt{F} \\ \hline
Middle MCP & \texttt{M} & \texttt{F} \\ \hline
Middle PIP & \texttt{M} & \texttt{F} \\ \hline
Ring MCP & \texttt{F} & \texttt{F} \\ \hline
Ring PIP & \texttt{F} & \texttt{F} \\ \hline
Pinky MCP & \texttt{F} & \texttt{F} \\ \hline
Pinky PIP & \texttt{F} & \texttt{F} \\ 
\end{tabular}
\end{table}

We consider each task as having to phases, a grasping phase, and a manipulation phase. For the grasping phase, the grasp synergy-based control method is used. For the manipulation phase, the proposed FDMS-based control method is used. To construct FDMS, it is necessary to define the function of the fingers for each task, scissors-task, and switch-task. \Cref{tab:selected_joints} shows the defined function of each finger when performing the scissors-task and the switch-task. These functions are empirically defined. The functions are assigned according to \Cref{ssec:listing_fdms}. The set consists of joint assigned \texttt{M}, and defined as $J^{\mathrm{sub}}$. We construct FDMS by using $J^{\mathrm{sub}}$ according to \Cref{ssec:synergy_definition}. Note that the maximum number of principal components is the number of joints assigned \texttt{M}. The switch-task has two phases: the grasping phase and the manipulation phase. Therefore, we switch synergies according to the Switching Synergy Framework, from grasp synergy to FDMS. 

\subsection{Evaluation}
\subsubsection{The task success rate}
In this experiment, we examined how many dimensions are reduced when using each synergy-based control method to perform the tasks in the real environment. The performance of conventional synergies was evaluated mainly by the contribution ratio which is percent variance accounted for by a principal component. However, it has not been verified whether tasks can be performed by using a synergy with a high contribution ratio.
To evaluate performance, we focus on the task success rate when replaying the posture sequences $\bm{X}^E_m(m=1\cdots 10)$ approximated by the synergy according to \Cref{ssec:postures_approximation} on the teleoperating system. The task success rate is calculated based on the percentage of successful tasks played by the approximated posture sequence.

\subsubsection{Practical demonstration}
We considered two more challenging tasks, a Cutting-task and a Spray-task. In the Cutting-task, the scissors placed on the workspace in advance are grasped, and a string is cut by operating the scissors. Similarly, in the Spray-task, the hand grasps and manipulates a spray bottle placed on the workspace by pulling the lever. We evaluate whether the Cutting-task and Spray-task can be realized, as more practical tasks. In this experiment, the trajectory of the hand is generated by our teleoperation system, but note that the posture obtained from CyberGloveI\hspace{-.1em}I\hspace{-.1em}I is approximated by our FDMS method in real-time. The pose of the object is known, and the movement of the arm is also determined in advance.

\section{results}
\label{sec:results}
\color{higashi}

\begin{figure}[t]
  \begin{center}
    \includegraphics[width=\linewidth]{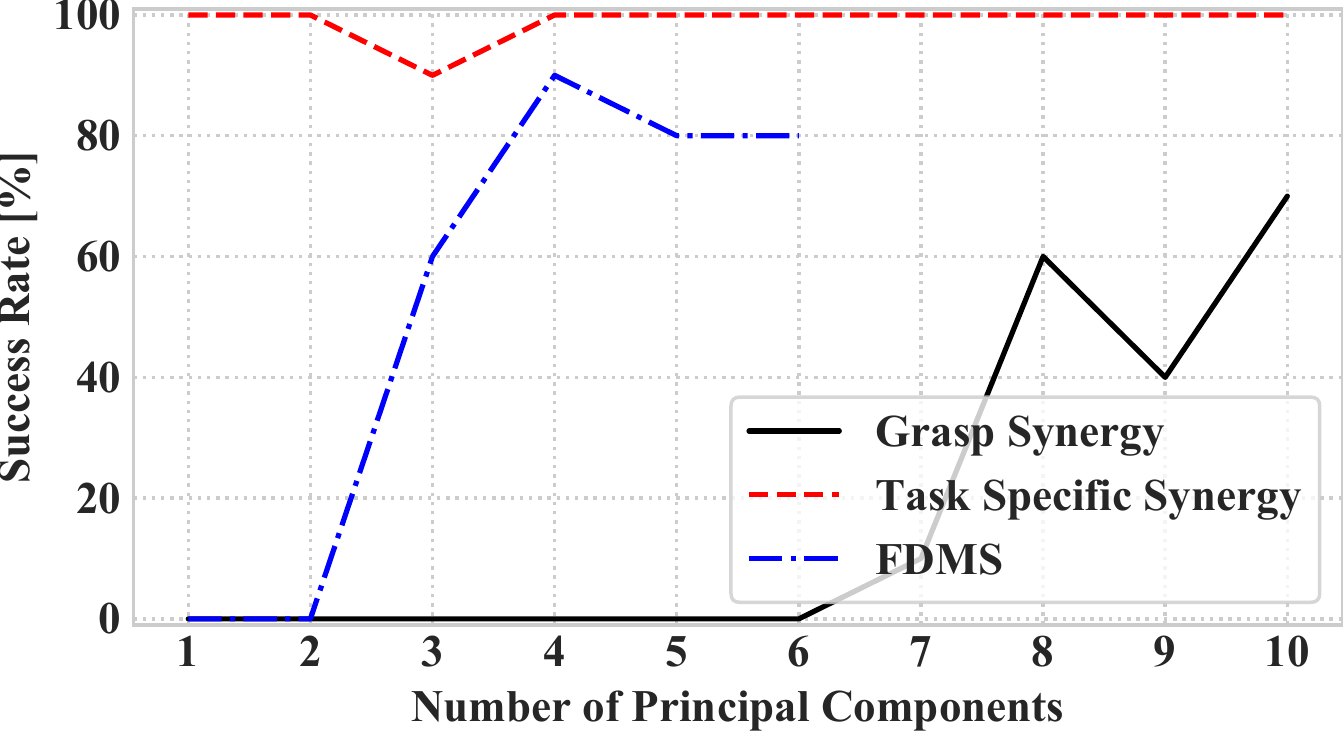}
    \caption{\color{higashi}\textbf{Scissors-task}: Relation between the number of principal components and success rates. The maximum number of principal components of FDMS for the scissors-task is 6 since the FDMS is composed 6 DoFs joint subspace.\color{black}} 
    \label{fig:scissor_result}
  \end{center}
\end{figure}

\begin{figure}[t]
  \begin{center}
    \includegraphics[width=\linewidth]{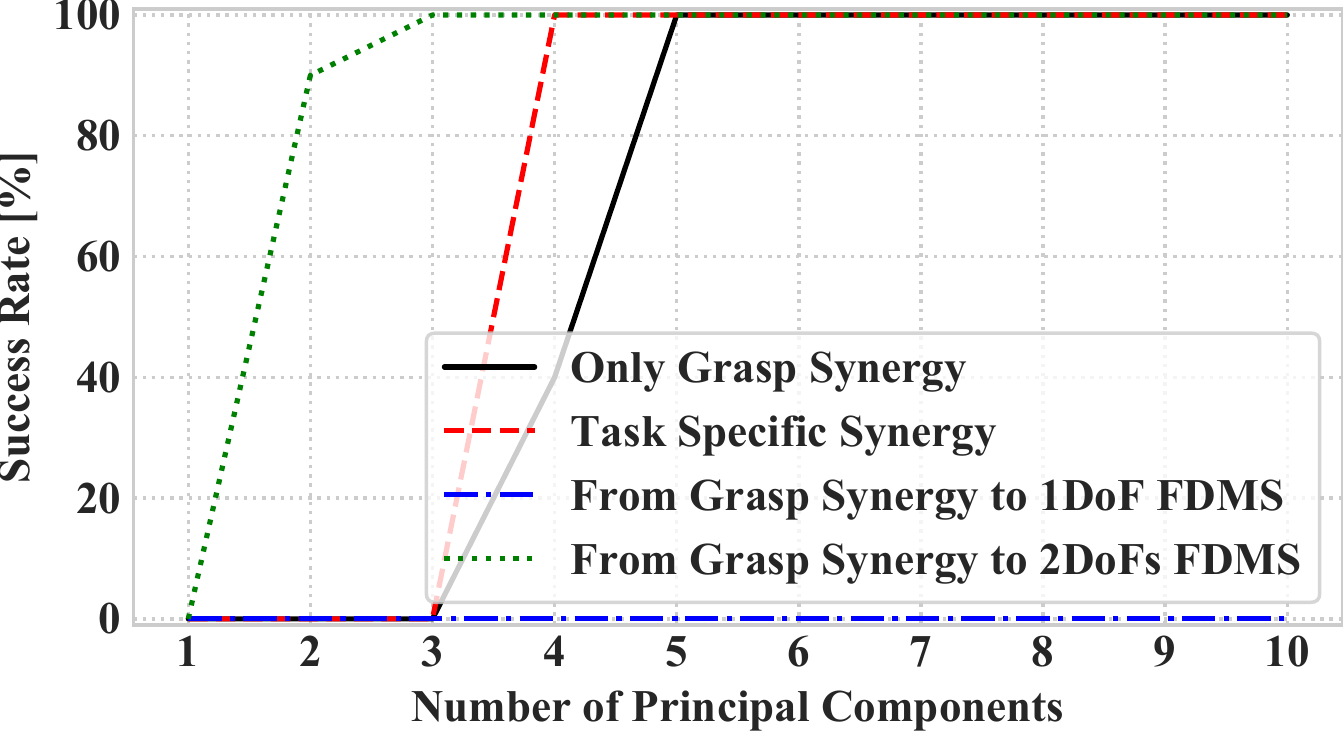}
    \caption{\color{higashi}\textbf{Switch-task}: Relation between the number of principal components and the success rates. Switch-task includes the change of synergy from grasp synergy to FDMS according to the Switching Synergy Framework since the fingers' function changes during the task. The blue and green lines show the success rate when the number of principal components of grasp synergy changes and we fix the number of principal components of the FDMS to 1 and 2 respectively.\color{black}}
    \label{fig:switch_result}
  \end{center}
\end{figure}

\color{higashi}
For each task, we show the plot of the success rate as a function of the number of principal components in \Cref{fig:scissor_result} (scissors-task) and \Cref{fig:switch_result} (switch-task). Both results show that, as the number of principal components increases, the task success rate increases. Regardless of the number of principal components, the success rate of using FDMS is higher than the success rate of using conventional grasp synergy. 

In the scissors-task, \Cref{fig:scissor_result}, the highest success rate when using the conventional grasp synergy is 70\%. In contrast, using our FDMS with four principal components achieved a task success rate of 90\%.
By fixing the position of some fingers, which are not necessary for the task, we reduce the overall DoF. Thus the accuracy of posture expression of FDMS improves due to the mathematical properties of PCA. On the contrary, if the position of those fingers is not fixed, then they may interfere with the manipulated object.
\color{cristian}

On the other hand, task-specific synergy achieved a task success rate of almost 100\% regardless of the number of principal components used. Therefore, when the number of tasks required for a humanoid hand is small, the said result implies that a high task success rate can be achieved by switching task-specific synergies according to the Switching Synergy Framework.

In the switch-task, \Cref{fig:switch_result}, when using the FDMS with just a single principal component, the task success rate is always 0 since the fingers whose function is the manipulating do not work well. This result shows that the thumb of the multi-fingered hand needs at least two DoFs. When using the grasp synergy with two principal components and the FDMS with two principal components, a task success rate of 90\% is achieved. In this case, the dimensions of the control space of the grasp synergy and the FDMS are always two, so it can be said that the effect of dimensionality reduction has been fully demonstrated. 

\section{discussion}
\label{sec:discussion}
\color{higashi}

As shown in \Cref{fig:scissor_result} and \Cref{fig:switch_result}, to achieve the task success rate of 100\% using the task-specific synergy, the scissors-task requires only one principal component, while the switch-task requires at least four principal components. The reason is that, in the switch-task, the fingers assigned for manipulating function changes from the index, middle, ring, and pinky fingers to the only thumb according to the transition of the task. Since the task-specific synergy assuming the coordination of all fingers, it usually cannot perform multiple functions. This result suggests that the Switching Synergy Framework, which can switch FDMS, is effective for tasks that are required to change the fingers' function.

\begin{figure}[t]
  \begin{center}
    \includegraphics[width=\linewidth]{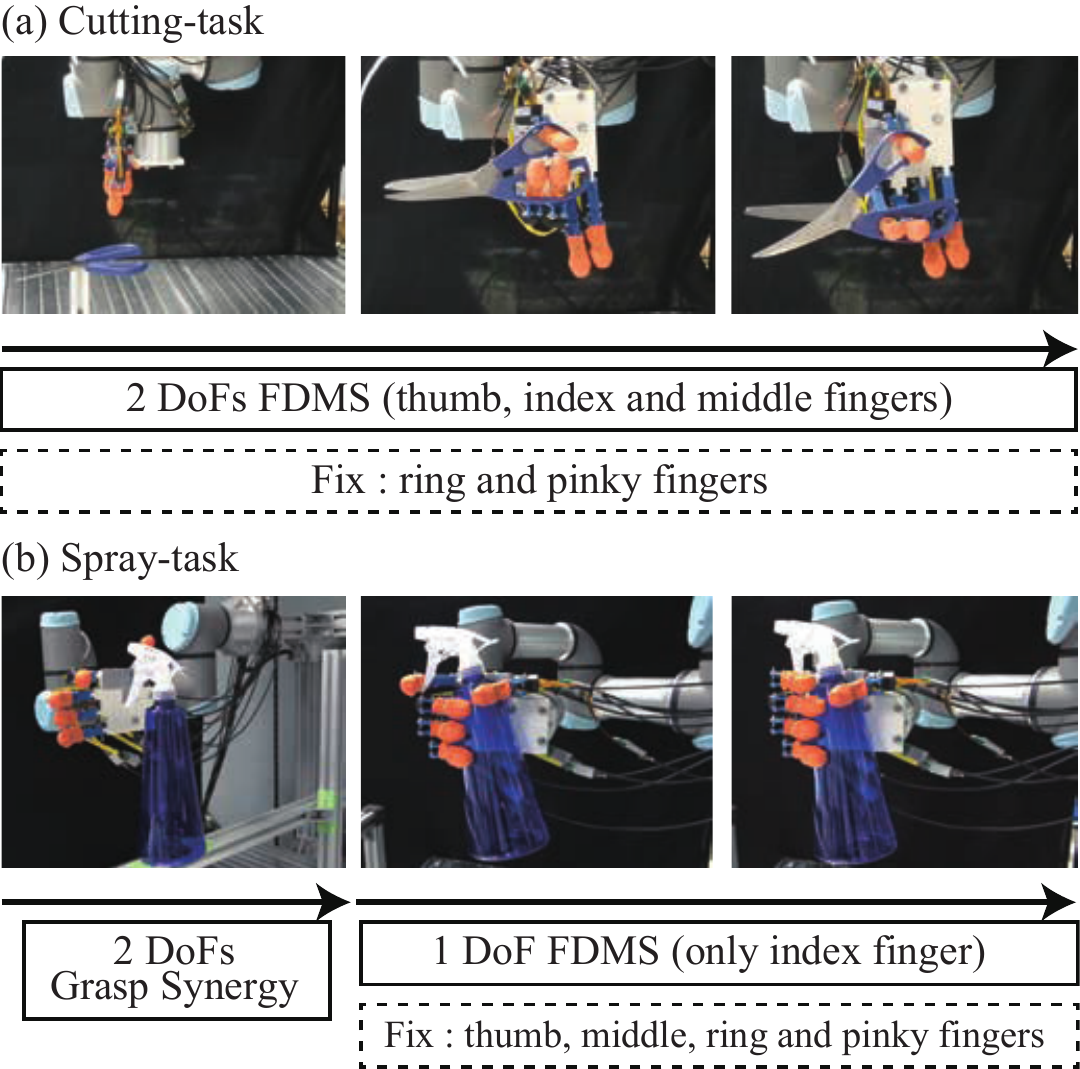}
    \caption{Demonstration of the cutting-task and the spray-task using FDMS-based control and the Switching Synergy Framework}
    \label{fig:fdms_demo}
  \end{center}
\end{figure}

\Cref{fig:fdms_demo} shows the hand performing the Cutting-task and Spray-task when using FDMS-based control and the Switching Synergy Framework. In the Cutting-task, the hand can grasp the scissors and cut the string by using only two principal components for the FDMS. In the spray-task, the hand has to grasp with all fingers and then push its lever with only the index finger while the other fingers maintain the grasping state. \color{higashi}Maintaining the grasp state is essential for manipulating the lever, this fact confirms the importance of fixing some of the fingers.\color{cristian} Thus, the grasp synergy-based control with a few principal components cannot realize the spray-task. In contrast, FDMS and the Switching Synergy Framework can realize the spray-task by using 2 DoFs grasp synergy and 1 DoF FDMS. 
These facts suggest that FDMS and Switching Synergy Framework effectively supply the control spaces spanned by a few principal components, which can realize various sequential tasks.

The conventional grasp synergy can generate trajectories to grasp objects \cite{prof.ficuciello1} and perform in-hand manipulation such as changing the pose of an object \cite{Pisa_SoftHand2}. Therefore, it is considered that FDMS, which is a form of grasp synergy limited by joint subspace, can generate the trajectory for grasping and manipulation.

In this study, we empirically determine the fingers' functions and its change during each task. However, in order to use FDMS and Switching Synergy Framework in a real environment, the functions required for a task should be automatically determined and assigned to the finger. In this study, it was possible to limit the number of FDMS required for manipulating an object to 12. Interesting future research would be to construct a Markov process with these 12 FDMSs as states. It is possible to construct a system that selects an appropriate FDMS from observation information of the operation target.

We define two functions to compose FDMS, but we can introduce more detailed functions to make our FDMS more useful. For instance, we can introduce a maintaining grasp function derived from the fixed function to control the internal force of a grasped object. 

\begin{table}[]
\centering
\caption{The number of principal components to achieve cumulative contribution ratio 80\% or 90\%}
\label{tab:variance_ratio_vs_PCs}
\begin{tabular}{c|c||r|r}
\multirow{2}{*}{Joint subspace} & \multirow{2}{*}{Movement unit} & \multicolumn{2}{c}{Number of PCs} \\ \cline{3-4} 
 &  & \multicolumn{1}{l|}{\textgreater{}80\%} & \textgreater{}90\% \\ \hline 
$J^{\mathrm{sub}}_1$ & \texttt{MMMMM} & 3 & 4 \\ 
$J^{\mathrm{sub}}_2$ & \texttt{MFFFF} & 1 & 2 \\ 
$J^{\mathrm{sub}}_3$ & \texttt{FMMMM} & 2 & 3 \\ 
$J^{\mathrm{sub}}_4$ & \texttt{MMFFF} & 2 & 3 \\ 
$J^{\mathrm{sub}}_5$ & \texttt{FFMMM} & 1 & 2 \\ 
$J^{\mathrm{sub}}_6$ & \texttt{FFFMM} & 1 & 2 \\ 
$J^{\mathrm{sub}}_7$ & \texttt{FMFFF} & 1 & 2 \\ 
$J^{\mathrm{sub}}_8$ & \texttt{MMMFF} & 3 & 4 \\ 
$J^{\mathrm{sub}}_9$ & \texttt{FFMFF} & 3 & 4 \\ 
$J^{\mathrm{sub}}_{10}$ & \texttt{FFMMF} & 1 & 2 \\ 
$J^{\mathrm{sub}}_{11}$ & \texttt{MMMMF} & 3 & 4 \\ 
$J^{\mathrm{sub}}_{12}$ & \texttt{FMMMF} & 2 & 3 \\ 
\end{tabular}
\end{table}

The posture expression ability required to realize a manipulating function is considered to be different for each FDMS. If we develop a multi-fingered hand that realizes FDMS, it is necessary to determine in advance how many principal components should be used for each synergy. Now, we discuss a method using a cumulative contribution ratio as one of the methods. Cumulative contribution ratio is the sum of contribution ratios from the first principal component to the $f$-th principal component in PCA and indicates how well the variance of the samples can be explained. Therefore, if an FDMS realizes a high cumulative contribution ratio, it is considered that the FDMS has a high posture expression ability. \Cref{tab:variance_ratio_vs_PCs} shows the number of principal components required for each FDMS shown in \Cref{tab:listup_fdsyn} to achieve a cumulative contribution ratio of 80\% or 90\%. The FDMS is the same as the one used in the experiment, defined in \Cref{ssec:exprimental_synergies}. Note that the maximum number of principal components of each FDMS is twice the number of \texttt{M} from the definition and the configuration of 10 DoFs multi-fingered hand. The results suggest that at most three and four principal components are sufficient to achieve the cumulative contribution ratios of 80\% and 90\%, respectively.
\section{conclusion}
\label{sec:conclusion}

In this paper, we proposed the Functionally Divided Software Synergy method, which can realize manipulation tasks using a multi-fingered hand, with a high task success rate. \color{higashi}It was confirmed that fixing some specific fingers was important for manipulating objects. Experiments confirmed that FDMS is effective for manipulation tasks.\color{cristian} Moreover, the required FDMS to manipulate an object can be limited to 12 based on the motion analysis of the human hand, and the cost is lower than constructing task-specific synergies, which need to configure synergies for each particular task.

Future works will aim to develop the method to design an under-actuated multi-fingered hand leveraging FDMS, and the system to assign functions to fingers according to information of an object.



\bibliographystyle{IEEEtran}
\bibliography{reference}

\end{document}